\documentclass{article}

\usepackage[final]{nips_2016}

\usepackage[utf8]{inputenc} 
\usepackage[T1]{fontenc}    
\usepackage{hyperref}       
\usepackage{url}            
\usepackage{booktabs}       
\usepackage{amsfonts}       
\usepackage{nicefrac}       
\usepackage{microtype}      
\usepackage{graphicx}
\usepackage{makecell}
\usepackage{algorithm}
\usepackage{algpseudocode}%
\usepackage{bm}
\usepackage{amsmath}

\title{Convolutional neural networks that teach microscopes how to image}

\author{
  Roarke Horstmeyer\thanks{Current address: NeuroCure Cluster of Excellence, Charit\'e--Universit{\"a}tsmedizin Berlin \& Humboldt University, Berlin Germany} \\
  Biomedical Engineering Department\\
  Duke University\\
  Durham, NC \\
  \texttt{roarke.horstmeyer@gmail.com} \\
  \And
 Richard Y. Chen   \hspace{1cm}\\
Y Combinator Research   \hspace{1cm}\\
San Francisco, CA    \hspace{1cm}\\
\texttt{richard.chen@ycr.org}  \hspace{1cm}\\
  \And
   \hspace{.7cm}Barbara Kappes \\
  \hspace{.7cm} Institute of Medical Biotechnology \\
   \hspace{.7cm}University of Erlangen-Nuremberg \\
   \hspace{.7cm}Erlangen, Germany \\
  \hspace{.7cm}\texttt{barbara.kappes@fau.de} \\
  \And
  Benjamin Judkewitz \\
  NeuroCure Cluster of Excellence \\
  Charit\'e--Universit{\"a}tsmedizin Berlin \& \\
  Humboldt University, Berlin, Germany\\
  \texttt{benjamin.judkewitz@charite.de} \\
}

\begin{document}

\maketitle

\begin{abstract}
Deep learning algorithms offer a powerful means to automatically analyze the content of medical images. However, many biological samples of interest are primarily transparent to visible light and contain features that are difficult to resolve with a standard optical microscope. Here, we use a convolutional neural network (CNN) not only to classify images, but also to optimize the physical layout of the imaging device itself. We increase the classification accuracy of a microscope's recorded images by merging an optical model of image formation into the pipeline of a CNN. The resulting network simultaneously determines an ideal illumination arrangement to highlight important sample features during image acquisition, along with a set of convolutional weights to classify the detected images post-capture. We demonstrate our joint optimization technique with an experimental microscope configuration that automatically identifies malaria-infected cells with 5-10\% higher accuracy than standard and alternative microscope lighting designs.    
\end{abstract}

\section{Introduction}

Artificial neural networks (ANNs) are now commonly used for image classification~\cite{Esteva17}, segmentation~\cite{Ronneberger15}, object detection and image resolution enhancement~\cite{Oord16}. Convolutional neural networks (CNNs) in particular continue to offer significant improvements over other supervised learning techniques. While the growing field of deep learning has seen a lot of progress over the past several years, most learning networks are trained on pre-captured datasets that contain standard images (e.g., MNIST~\cite{LeCun98}, CIFAR-10, or BBBC~\cite{Ljosa12}). Little work with ANNs and CNNs has jointly considered \emph{how} the image data that they process is actually acquired - that is, whether a particular camera setup, illumination geometry or detector design might improve certain image post-processing goals. 

In many scenarios, and in optical microscopy in particular, the various physical parameters used to obtain an image are extremely important. For example, one must select an appropriate microscope objective lens that fits the entire area of interest within its field-of-view, but also still resolves important features of interest. Or, primarily transparent samples are often hardly visible at all under a standard bright-field microscope. Alternative techniques like phase-contrast or differential interference contrast microscopy are typically used improve sample visibility. Likewise, more recent computational approaches can also create sharper images by using structured illumination~\cite{Gustafsson00, Betzig06, Zheng13}. However, it is not clear which type of microscope design, illumination setting, or computational optics-based approach might yield the best results for a particular CNN-driven task.

Here, we try to close the gap between how images are acquired and how they are post-processed by CNNs. We include a model of optical image formation into the first several layers of a neural network, which allows us to jointly optimize the design of a microscope and the various weights used for image classification in one step, thus forming a specific type of ``classification microscope." As an experimental demonstration, we jointly determine an optimal lighting pattern for red blood cell imaging as well as a CNN classifier to test for infection by the malaria parasite ({\it Plasmodium falciparum}). By simply displaying two particular patterns on an LED array placed beneath our microscope, we can increase infection classification accuracy by 5-10\%. We are hopeful that the presented framework will help bring together the growing field of deep learning with those who design the cameras, microscopes and other imaging systems that capture the training data that most learning networks currently use.

\begin{figure}
\centering
\includegraphics[width=0.85\linewidth]{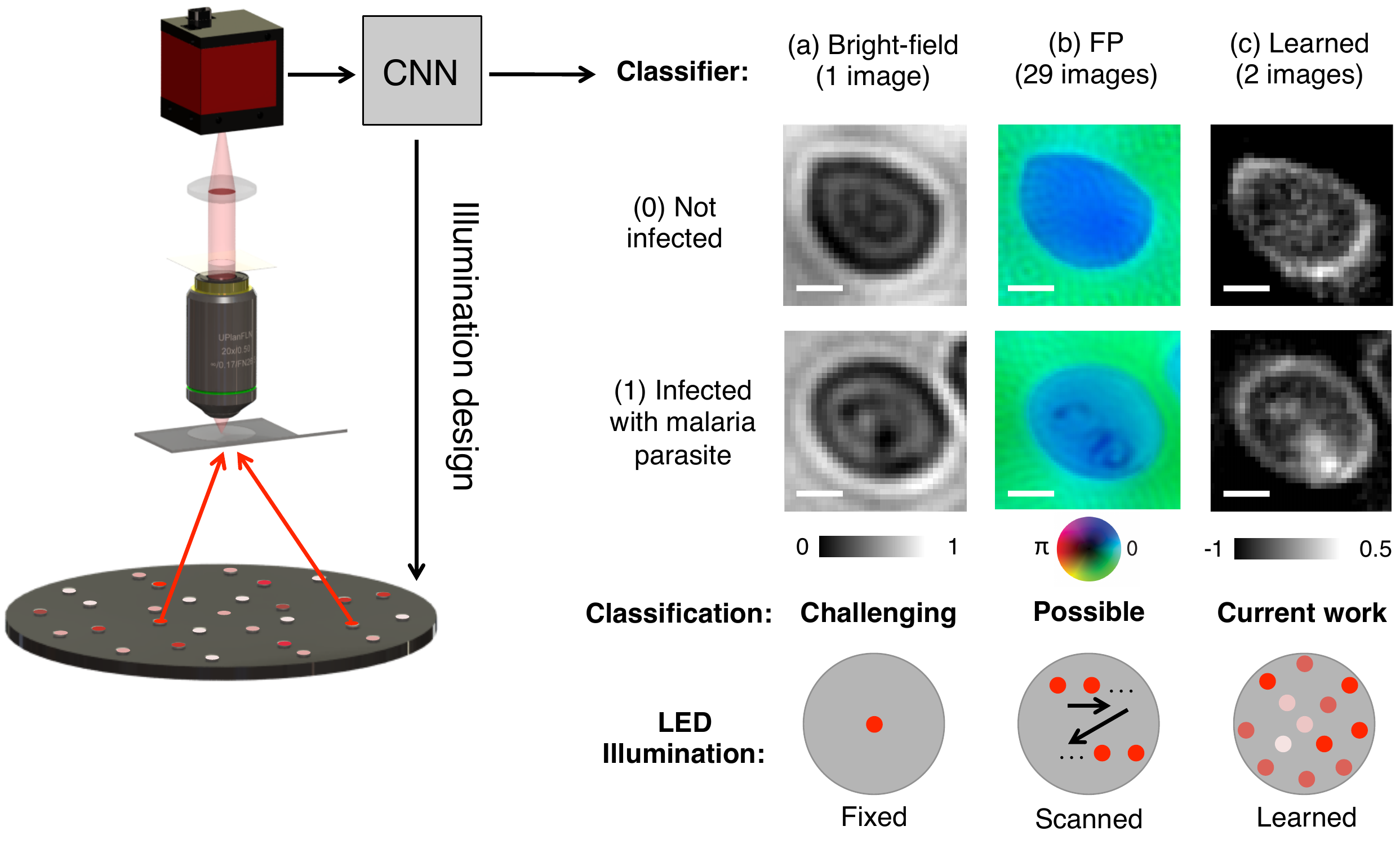}
\caption{We use a convolutional neural network to jointly optimize the physical layout of microscope illumination as well as a classifier to determine if cells are infected by the malaria parasite. The classification accuracy with images from our optimal illumination technique in (c) notably exceeds that for a standard bright-field imaging setup in (a). Alternative schemes like Fourier ptychography (FP, complex-valued reconstruction in b) can improve spatial resolution for better subsequent classification, but must capture more images and are thus less efficient. Scale bar: 2 $\mu$m.}
\label{teaser}
\end{figure}


\section{Related work}

Deep learning networks for conventional~\cite{Lecun15} and biomedical~\cite{Litjens17} image classification are now commonplace and much of this recent work relies on CNNs. However, as noted above, most studies in this area do not try to use deep learning to optimize the acquisition process for their image data sets. While early work has shown that simple neural networks offer an effective way to design cameras~\cite{Macdonald93}, one of the first works to consider this question in the context of CNN-based learning was presented recently by Chakrabarti~\cite{Chakrabarti16}, who designed an optimal pixel-level color filter layout for color image reconstruction that outperformed the standard Bayer filter pattern.

Subsequent work has also considered using CNNs to overcome the effects of camera sensor noise~\cite{Diamond17} as well as optical scattering \cite{Horisaki16, Satat17}. CNNs paired with non-conventional camera architectures can additionally achieve light-field imaging~\cite{Chen16, Wang17} and compressive image measurement~\cite{Kulkarni16}. Alternative imaging setups have also been coupled with supervised learning methods for cell imaging and classification~\cite{Park16, Huang16}. The above works are some of the first to consider how the performance of supervised learning in general and CNNs in particular connect to the exact procedure for optical data acquisition. But, with the except of Chakrabarti's work, they do not directly consider using the CNN itself to optimize the detection process of an optical (or digital) device. 

In this work, we first merge a general model of optical image formation into the first layers of a CNN. After presenting our "physical CNN" model, we then present simulations and experimental results for improved classification of cells from microscopic images. We focus on the particular task of classifying if red blood cells are infected with the {\it Plasmodium falciparum} parasite, which is the most clinically relevant and deadly cause of malaria. This goal has been examined within the context of machine learning~\cite{Park16,Das13} and CNNs~\cite{Quinn16} in the past. Unlike prior work, we demonstrate how our physical CNN can simultaneously predict an optimal way to illuminate each red blood cell. When used to capture experimental images, this optimized illumination pattern produces better classification scores than tested alternatives, which hopefully can be integrated into future malaria diagnostic tools.

\begin{figure}
\centering
\includegraphics[width=0.77\linewidth]{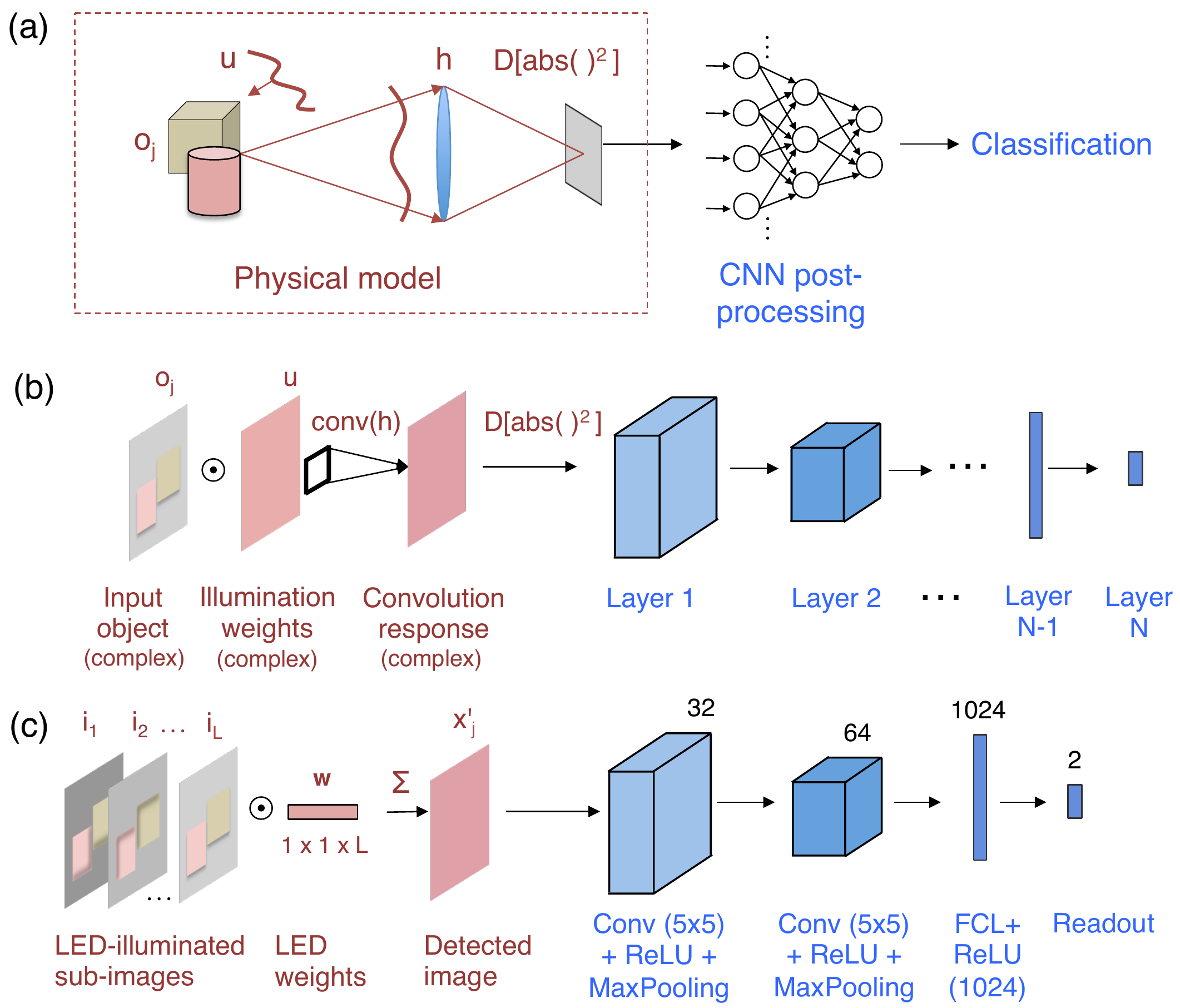}
\caption{Physical CNN (P-CNN) for improving image classification accuracy. (a) Standard imaging setup with variables of interest labeled, followed by a digital CNN. (b) P-CNN pipeline. Physical layers (here including illumination, PSF convolution and detection) are in red while post-processing layers are in blue. (c) Discretized physical CNN (DP-CNN) network used in simulations and experiments. Each sub-image, under illumination from a unique direction, is weighted and summed to form the final detected image that then enters a CNN classifier. Learned LED weights from the physical layer specify the optimal illumination design for our classification microscope. }
\label{cnnmodel}
\end{figure}


\section{Image formation and network layers}

In a typical imaging scenario, light first travels from a source to a scene of interest, physically interacts with objects within the scene, and then propagates through a lens and to a sensor to form an image. Here, we will model these physical processes in the early ``physical" layers of a neural network, which will then be followed by a series of standard "post-processing" network layers. In this work we design the post-processing layers for image classification, but they could also be easily guided towards other standard tasks (e.g., segmentation or object detection). After network optimization, the physical-model weights, associated with the experimental parameters used to capture each image, can then directly inform an optimized optical device design. 

In a standard image-based CNN, inputs take the form of pre-recorded incoherent intensity images. Each input image $x_j$, for $j=1$ to $J$ training images, then passes through a number of network layers. In each early network layer, 3 simple operations are typically applied in a repeated manner (e.g.,as used in the early 5-layer LeNet~\cite{LeCun98}, the more recent AlexNet~\cite{Krizhevsky12}, and many others since). The first operation is a convolution, the second is a non-linear activation function $L$, and the third is a down-sampling operation $D$. We may write these three steps as,
\begin{equation}
x'_j = D\left[ L\left[ x_j \star h\right] \right],
\label{onelayer}
\end{equation}   
where $x'_j$ is the layer output, $h$ is the convolution kernel and $\star$ denotes convolution. A common choice for the non-linear operation $L$ is a rectified linear unit (ReLU) and a common choice for the downsampling step $D$ is max-pooling. 

Here, we take a step back and first consider how light behaves before forming an image. Let's consider the simplified imaging system in Fig.~\ref{cnnmodel}(a) that we assume collects images for a classification task. In this simple scenario, let's assume that the object of interest is thin, which allows us to model its optical response with a complex vector $o_j$. Furthermore, we'll assume that our object is illuminated with spatially coherent light source (e.g., from a small source like an LED, as is often the case in a microscope, although it is direct to model any other type of light). If we label our illumination field $u$, then we can approximate the resulting optical field that is reflected off and/or transmitted through the thin object as an element-wise multiplication between two complex vectors:  $x_j = u \cdot o_j$.

After emerging from the object, the optical field $x_j$ will then propagate to the imaging system (e.g., the camera lens in Fig.~\ref{cnnmodel}(a) or a microscope objective lens). A portion of this field will pass through the lens and eventually reach a digital detector comprised of an array of pixels, which detects and discretizes the intensity of the field. It is very common to model the linear process of image formation as a convolution of the optical field at the object plane with a camera-specific point-spread function (PSF), $h$~\cite{Goodman}. The detector then measures the squared magnitude of the resulting convolved field. If we summarize the discretization (i.e., locally averaging) caused by the finite size of each pixel with a pixel sampling function $P$, then the entire image formation process takes the following form:
\begin{equation}
x'_j = P\left[ \left| x_j \star h\right|^2 \right].
\label{coherentcamera}
\end{equation}      
Here, $x_j=u \cdot o_j$ is the product of the illumination light with the object and $x'_j$ is the resulting image. For a square pixel array with a full fill-factor, $P$ is approximately equivalent to an average pooling operation. Comparing Eq.~\ref{onelayer} with Eq.~\ref{coherentcamera}, we see that image formation is nearly identical to an early CNN layer. The two models are equivalent when 1) the CNN layer's non-linear activation function $L$ is an absolute value squaring such that $\forall x$, $L(x)=|x|^2$ (equivalent to a leaky ReLU function with parameter $\alpha=-1$) and 2) the downsampling operation $D$ is average pooling. 

Given the similarity between Eq.~\ref{onelayer} and Eq.~\ref{coherentcamera}, we hypothesize that Eq.~\ref{coherentcamera} is a natural fit for the first layer of a CNN whose inputs are representations of physical objects (as opposed to their images). While the weights within subsequent layers may be dedicated to computational operations for a particular classification task, weights within this early "physical model" layer directly connect to experimental parameters like the illumination field $u$, the camera's point-spread function $h$ and the downsampling function $P$.  After network optimization, they may inform a better imaging setup and measurement process to achieve a particular CNN-related task. In the next section, we will focus our attention on optimizing the illumination field $u$ to improve image classification, but this framework generalizes to optimizing both other optical parameters in Eq.~\ref{coherentcamera} ($h$ and $P$) and other optical parameters of interest (e.g., illumination wavelength, polarization, camera/lens location).

\section{Optimal illumination for classification}

Here, we first explain the P-CNN model in Fig.~\ref{cnnmodel}(b) in more detail and then discuss the particular network used in our experiments (Fig.~\ref{cnnmodel}(c)). Let's assume the $j$th input into our P-CNN is a complex $N\times N$ matrix $o_j$ describing the $j$th sample of interest (e.g., a cell or collection of cells under a microscope that we want to classify). Under the thin sample approximation, the sample absorption is represented by its modulus $\left|o_j\right|$ and the sample phase response, which is typically directly proportional to its thickness, is represented by its argument $\arg\left(o_j\right)$. The coherent illumination field $u$ is modeled as an $N\times N$ complex matrix of unknown illumination weights (assuming it can take any desired shape), and as noted above the field emerging from the sample, $x_j$ in Eq.~\ref{coherentcamera}, is an element-wise product of $o_j$ and $u$. To transform $x_j$ into a detected image, we then perform a convolution with a complex point-spread function $h$ (1 filter, stride = 1) and an element-wise vector squaring and down-sampling, which simulates $x'_j$ in the P-CNN's first physical layer (Fig.~\ref{cnnmodel}(b), red). Any standard convolution-based classification network (Fig.~\ref{cnnmodel}(b), blue) can follow this first layer. 

After network optimization, the learned $N \times N$ weights in $u$ specify the optimal magnitude and phase of an optical field to illuminate a sample with for subsequent classification. Since patterned illumination can provide better image contrast, highlight features of interest and shift higher spatial frequencies into a camera's finite frequency passband, we hypothesize that images formed under optimal physical illumination can be classified more accurately by the P-CNN post-processing layers than those formed by alternative illumination strategies. 

\begin{figure}
\centering
\includegraphics[width=0.8\linewidth]{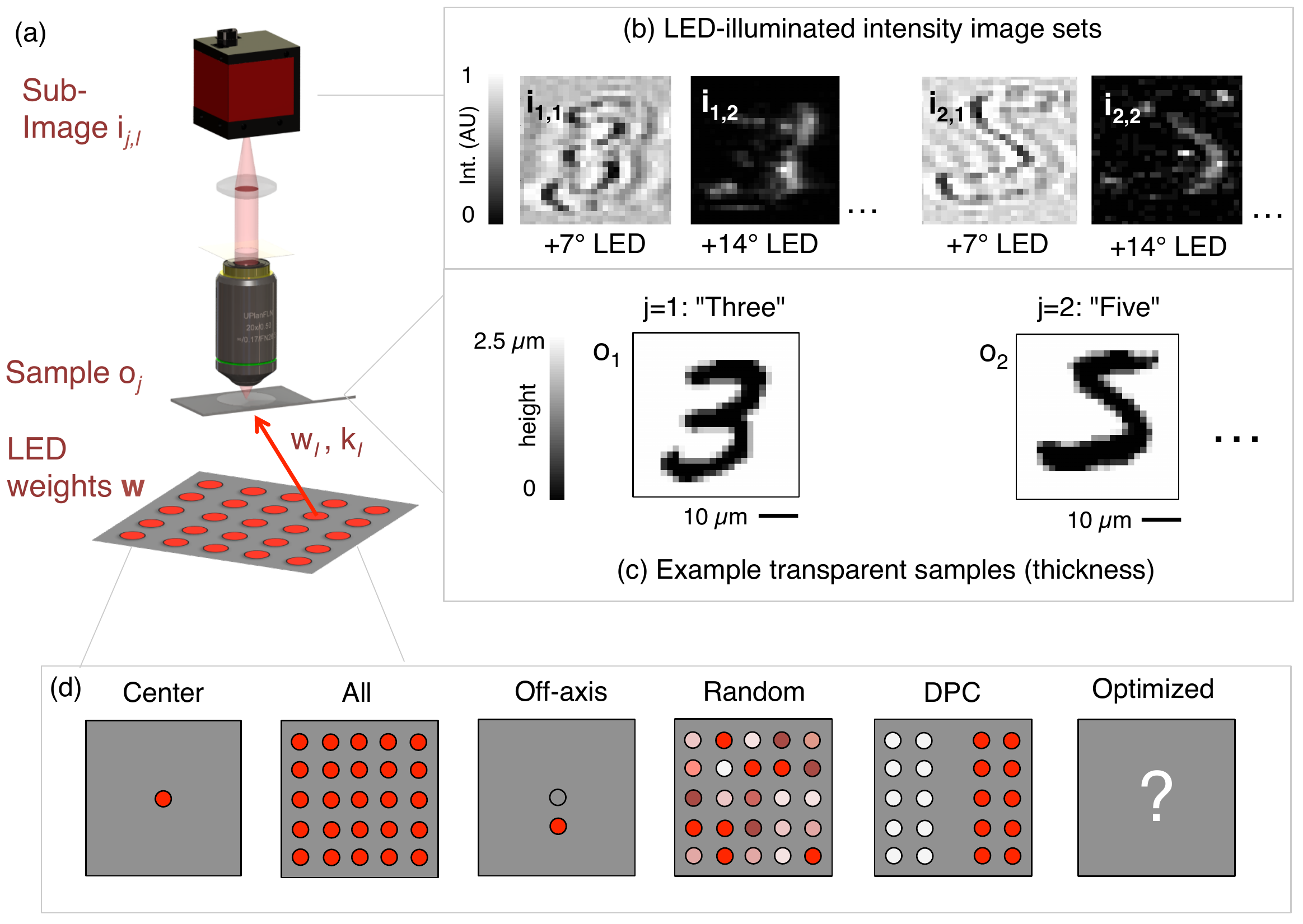}
\caption{(a) Simulated microscope setup. (b) 2 out of 25 example intensity images are displayed for 2 unique image sets. Each image (per set) is illuminated from a unique LED angle. (c) Example transparent objects created from the MNIST dataset and used to generate images in (b). (d) Illumination patterns used for classification tests (optimized pattern determined via DP-CNN learning).}
\label{simsetup}
\end{figure}


While in practice one could physically shape the optimal field $u$ using an $N \times N$ pixel spatial light modulator, we instead choose to work with the alternative imaging setup sketched in Fig.~\ref{simsetup}(a). Here, a discrete set of LEDs from an array define our sample illumination. Each LED in the array illuminates the sample with a coherent plane wave from a unique angle. Since the LEDs are mutually incoherent with one another, the resulting image from multiple LEDs emitting simultaneously is the incoherent sum of each LED-illuminated ``sub-image". If we have $L$ LEDs within the array, each emitting a field with magnitude $\sqrt{w_l}$ for $l=1$ to $L$, then we may write the $j$th detected image as the incoherent sum,
\begin{equation}
x'_j=\sum_{l=1}^L \left| (o_j \cdot \sqrt{w_l} e^{i\bf{k_l} \cdot \bf{r}}) \star h \right|^2=\sum_{l=1}^L w_l i_{j, l}.
\label{sumintensity}
\end{equation}
Here, $\sqrt{w_l} e^{i\bf{k_l} \cdot \bf{r}}$ describes the $l$th LED-generated plane wave with intensity $w_l$, traveling at an angle specified by the transverse wavevector $\bf{k_l}$ with respect to the optical axis. After each plane wave illuminates the object $o_j$, it is then blurred by the imaging system (coherent point-spread function $h$) to form the $l$th sub-image of the $j$th object, $i_{j, l}$. The weighted sum of these sub-images, where the weight $w_l$ specifies the $l$th LED brightness, forms the detected image under illumination from a particular LED pattern, $x'_j$. This image then enters a digital classification pipeline. We refer to this as the discretized physical CNN (DP-CNN) architecture, summarized in Algorithm~\ref{algo:DP-CNN}. 

For training, the DP-CNN simply requires the set $\{i_{j, l}\}_{l=1}^L$ of $L$ uniquely illuminated images of each object of interest as input. After optimizing the LED weight vector $w$ for a particular classification task, its values then specify how to \emph{simultaneously} illuminate each future sample of interest with the LED array for improved classification. Each single optimally illuminated image will then enter the post-processing classification layers of the trained DP-CNN pipeline (blue layers in Fix.~\ref{cnnmodel}(c)), since the red layers of the DP-CNN have already been physically realized during image recording. By illuminating the sample at a wide range of angles, we are able to physically shift around the sample spectrum (i.e., the Fourier transform of the sample function $o_j$) to capture a wide range of its spatial frequencies and thus a lot of diverse information that would otherwise not enter the camera or microscope. In simulation and experiment, we illuminate from angles that are both less than and greater than the maximum acceptance angle of our microscope objective lens to form both bright-field and dark-field images of each sample. As we will show next, the classification accuracy for DP-CNN-optimized images exceeds alternative illumination approaches.

\begin{algorithm}[t]
\caption{Discretized Physical CNN (DP-CNN)}
\label{algo:DP-CNN}
\begin{algorithmic}[1]
\State {\bf Input}:  $\{i_{j, l}\}$ with $l \in \{1, \dots, L\}$ uniquely illuminated images of each object, object labels $\{y_j\}$. Number of iterations $T$. An image classification model $F$ parametrized by $\bm{\theta}$. 
\State Randomly initialize the LED weights $\bm{w} = \{w_l\}_{l=1}^L$. 
\For{iteration $t = 1 \dots, T$}
	\State Sample a minibatch of $( \{i_{j, l}\}_{l=1}^L, y_j)$,
	\State Generate the detected images $\{x_j'\}$ according to Eq.~\ref{sumintensity},
	\State Take a gradient descent step on $\mathrm{CrossEntropy}(F(x_j'), y_j)$ with respect to $\bm{w}$ and $\bm{\theta}$.
\EndFor
\State {\bf Output:} the optimized LED weights $\bm{w}$ and the model parameters $\bm{\theta}$.
\end{algorithmic}
\end{algorithm}

\section{Simulation results}

To test whether the DP-CNN leads to an improved classification microscope design, we first simulate a realistic imaging scenario. Many microscopic samples of interest are primarily transparent and are thus challenging to capture clear images of under standard illumination. The first step in our simulation is to generate many different transparent ``samples" that we would like classify. To do so, we use a standard classification dataset, the MNIST handwritten digit image set~\cite{LeCun98}, and treat each normalized MNIST image $g_j(x,y)$ as the height profile map of a thin, optically clear object. Conceptually, this is equivalent to assuming that instead of being written in ink, each digit was pressed into a clear plastic sheet. 

Under the thin object approximation, we may express the phase delay of normal incident light that passes through each of these transparent digit sheets as $\Delta\phi_j(x,y)=2\pi n t g_j(x,y)/\lambda$, where $n$ is the medium's refractive index, $t$ is the maximum height variation, and $\lambda=500$ nm is the wavelength of quasi-monochromatic illumination. We also make the realistic assumption that the sample is not 100\% transparent by defining a small amount of absorption as $A_j(x,y) = \alpha g_j(x,y)$ with constant $\alpha=0.01$. The optical field that would emerge from the sample under flat plane wave illumination, $o_j(x,y)$, is then expressed as the complex function $o_j(x,y)=A_j(x,y)\mathrm{exp}(i\Delta\phi_j(x,y))$. 

\begin{table}[]
  \caption{Simulation results, MNIST phase digit classification with DP-CNN}
  \label{sample-table}
  \centering
  \begin{tabular}{llllllll}
    \toprule
    \cmidrule{1-2}
    Noise Level & Center & All & Off-axis & Random & DPC & Optimized   \\
    \midrule
    None ($\sigma$=0) & 0.378 & 0.975 & 0.974 & 0.975 & 0.974 & {\bf 0.983}   \\
    Low ($\sigma$=0.025) & 0.362 & 0.973 & 0.974 & 0.972 & 0.974 & {\bf 0.983}   \\
    High ($\sigma$=0.1) & 0.367 & 0.971 & 0.967 & 0.970 & 0.972 & {\bf 0.981}   \\
    \midrule
    Average & 0.369 & 0.973 & 0.972 & 0.972 & 0.973 & {\bf 0.982} \\
    \bottomrule
  \end{tabular}
  \label{simtable}
\end{table}

It is possible to directly use the sample responses $o_j(x,y)$ and their associated labels $y_j \in$ (0--9) as training/testing inputs for the P-CNN in Fig.~\ref{cnnmodel}(b). To test the DP-CNN in Fig.~\ref{cnnmodel}(c), we must first simulate uniquely illuminated images of each object with Eq.~\ref{sumintensity}. For each of $j=1$ to 60,000 transparent digits described by $o_j$, we simulate 25 different microscope images, $i_{j, l}$,  by turning on each of the $l=1$ to 25 LEDs in a $5\times5$ square array located beneath the sample plane (one LED at a time). We assume each sample has an index of refraction $n=1.2$, is at a maximum $t=2.5$ $\mu$m thick, is $40 \times 40$ $\mu$m in size and is discretized into 1.25 $\mu$m pixels. We use a 5X microscope objective with a 0.175 numerical aperture (NA) for simulated imaging and we illuminate each transparent digit sample with the $5\times 5$ LED array placed 50 mm beneath the sample for uniform angular illumination at 0, 7.2 and 14.5 degrees (on-axis, 1st off-axis and 2nd off-axis LEDs, respectively). The image detector contains $N^2=28\times28$ pixels that are each 7 $\mu$m in width and exhibits 1\% Gaussian readout noise. The height profile of two example transparent digits and their resulting images are in Fig.~\ref{simsetup}(b)-(c). 

We split the set of $J\times L = 60,000\times25$ classified digit images into a training set ($50{,}000 \times 25$ images) and testing set ($10{,}000 \times 25$ images) and run the 5-layer DP-CNN in Fig.~\ref{cnnmodel}(c) (using Tensorflow, see Appendix A for algorithm details). We add a different amount of random Gaussian noise at the sample plane with a scaling parameter $\sigma$ to test the robustness of our physical weights. We independently train and test the system 5 times for 3 different levels of noise and report the mean performance in Table \ref{simtable}, where we compare the performance of the full DP-CNN pipeline with several other illuminate-and-classify strategies sketched in Fig.~\ref{simsetup}(d). When simulating alternative LED patterns for sample illumination, we fix $w$ as a constant and do not attempt to learn the illumination weights (only the post-processing layers are optimized, not the physical layer). When we include the physical layer in the ``optimized" case, DP-CNN training produces an optimized LED pattern that we then use to illuminate the sample during network testing. 

\begin{figure}
\centering
\includegraphics[width=0.75\linewidth]{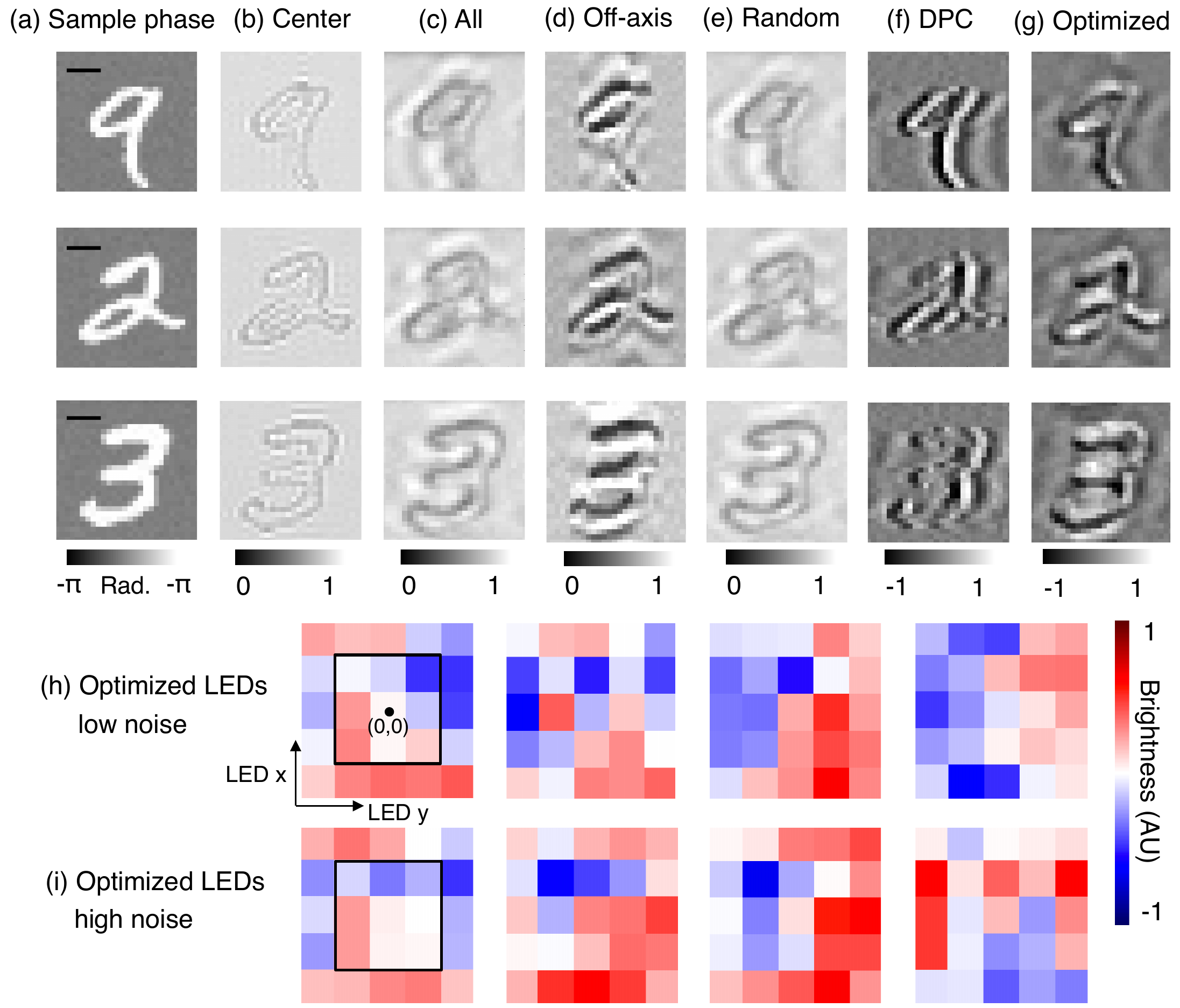}
\caption{Simulation results. (a) Example phase maps of transparent samples (scale bar: 10 $\mu$m). (b)-(g) Associated images under the different forms of LED illumination sketched in Fig.~\ref{simsetup}(d). Images in (b)-(d) are from one image capture, while images in (e)-(g) are from two image captures followed by a subtraction to allow for LED negative weights. (h)-(i) Example optimized LED pattern weights for the $5\times5$ LED array used to generate optimized images as in (g).  }
\label{simresults}
\end{figure}

Illuminating with only the center LED produces low-contrast images (spatially coherent bright field illumination, Fig.~\ref{simresults}(b)). Classification using this pattern is thus extremely poor: 36.9\% on average. Turning on an off-axis LED produces an image with phase contrast (Fig.~\ref{simresults}(d)) that significantly boosts classification accuracy to 97.2\%. So does illuminating with all of the LEDs simultaneously (spatially incoherent Kohler-type illumination, Fig.~\ref{simresults}(c), 97.3\%). 

The above three illumination schemes assume that each LED weight is positive. We can also test strategies where the LED weights are either positive or negative. This can be achieved in practice by capturing two images - a first with the positively weighted LEDs on, and a second with the negatively weighted LEDs on - and then digitally subtracting the second image from the first. We test this two-image strategy in simulation for randomly weighted LEDs (Fig.~\ref{simresults}(e), 97.2\% average), a differential phase-contrast (DPC) configuration~\cite{Mehta09} where the right half of the LED array has uniformly positive weights and the left half has uniformly negative weights (Fig.~\ref{simresults}(f), 97.3\% average), and finally our DP-CNN optimization pipeline (Fig.~\ref{simresults}(g), 98.2\% average). From the example images, it is clear that the optimal illumination strategy produces both a large amount of contrast as well as a relatively sharp image of each transparent digit, which leads to the highest classification accuracy across all tested noise levels. Finally, we note that the post-processing layers of our DP-CNN yield 99.3\% classification accuracy when trained and tested with an unmodified MNIST image set. So, at least in this example, phase-only samples are more challenging to classify than absorptive samples (that is, the relatively unclear images in Fig.~\ref{simresults} are harder to classify than the original MNIST set).

An example collection of DP-CNN optimized ``digit classification" LED patterns is in Fig.~\ref{simresults}(h)-(i). The black box at left marks the cutoff between bright field and dark-field LEDs (that is, LEDs at an angle less than and greater than the lens acceptance angle). Noting positive weights are in red and negative are in blue, ideal illumination strategies appear to approximately involve subtracting two brightly-lit images from two different and opposing directions within the bright-field channel, as is used for DPC imaging. This makes intuitive sense - since the relevant information needed to classify each digit resides in its phase, the optimal illumination strategy should maximize the amount of phase contrast within the detected images. However, the opposing bright LEDs do not form a symmetric pattern, and the optimization process clearly does not converge to a global minimum in this simulation.

\section{Experimental results}

Next we test the DP-CNN pipeline in a practical image classification experiment: to automatically determine if red blood cells are infected with the {\it P. falciparum} parasite, a primary cause of malaria. The infection typically manifests itself as a small spot or ring inside the red blood cell body. Currently, the leading method for diagnosing {\it P. falciparum} infection is manual inspection of stained blood smears by a clinical expert with an oil immersion microscope~\cite{WHO09}. To design a classification microscope that can automatically and accurately detect infection via our new illumination strategy, we outfit a standard Olympus microscope with an LED array positioned 40 mm beneath the sample~\cite{Ou15, Horstmeyer15}. We use 29 different LEDs (model SMD 3528, 20 nm spectral bandwidth, 150 $\mu$m active area diameter) that are arranged in 4 concentric rings. We image each sample across a wide field-of-view (FOV) using a 20X objective lens (0.5 NA, Olympus UPLFLN, 30 degree collection angle) and a large format CCD detector (Kodak KAI-29050, 5.5 $\mu$m pixels). The maximum angle of illumination from the LED array is approximately 45 degrees (0.7 illumination NA). Under high-angle illumination, we can thus capture spatial frequencies that are as large as the maximum frequency cutoff of a 1.2 NA oil immersion objective lens~\cite{Ou15}. By using a 20X objective lens instead of a high magnification oil immersion lens, we can image many more cells within its larger FOV but at the expense of a lower image resolution. We hope to overcome the limited 20X image resolution with our optimized illumination strategy, which will allow us to simultaneously offer accurate classification and high detection throughput in an oil-free imaging setup.  

Our samples are a number of different slides of single-layer human blood smears. Some of the red blood cells within each slide are infected with {\it P. falciparum} while the majority of cells are not infected. Of the infected cells, the majority are infected with {\it P. falciparum} at ring stage. A minority of infected cells are trophozoites and schizonts, which help test how well the DP-CNN classifier can generalize to different stages of infection. Each slide has been lightly stained using a Hema 3 stain. 

\begin{figure}
\centering
\includegraphics[width=0.8\linewidth]{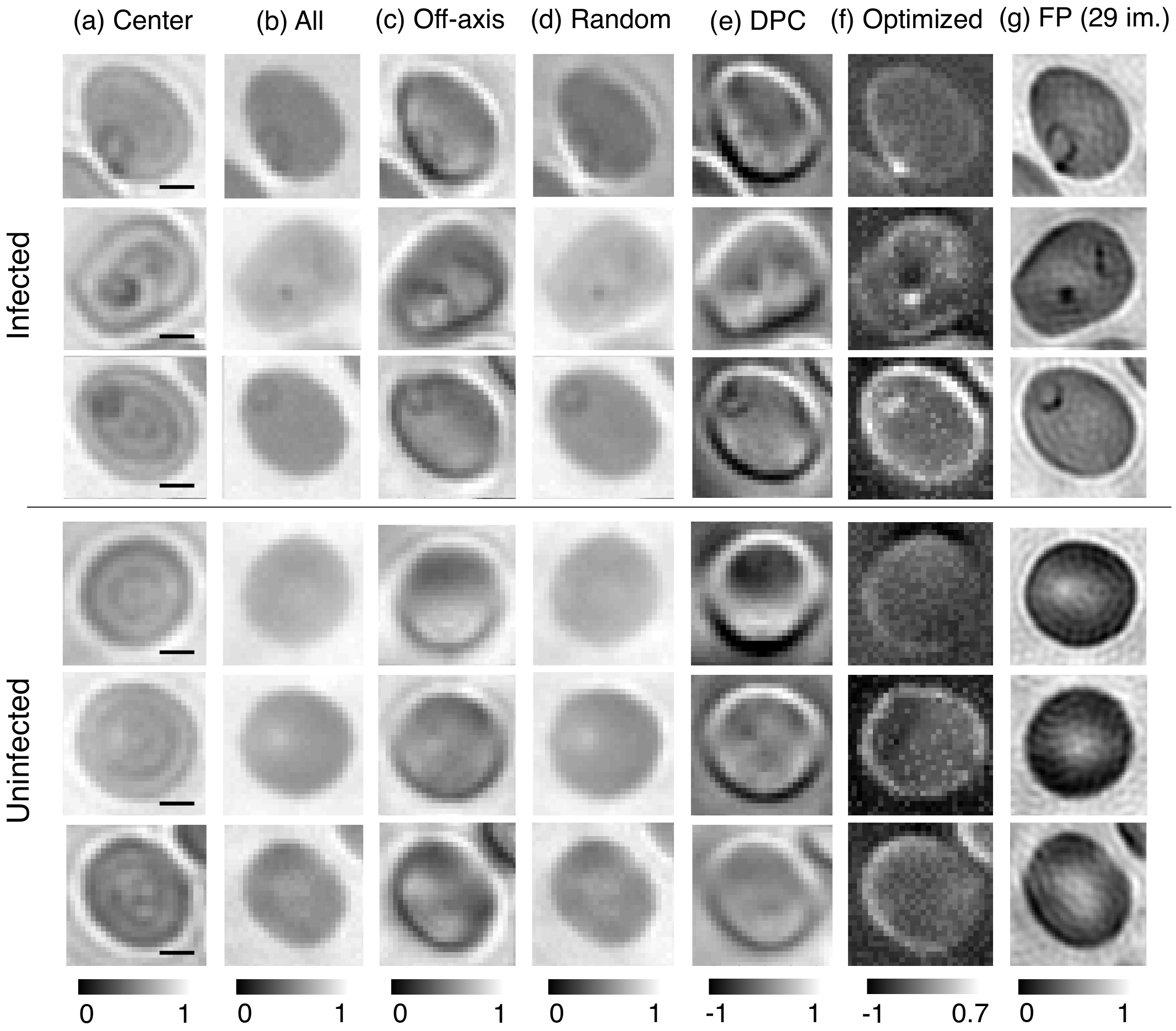}
\caption{Experimental results of {\it P. falciparum}-infected cells (top 3 rows) and uninfected cells (bottom 3 rows) using different LED illumination patterns (scale bar: 2 $\mu$m). (a)-(c) Images from one image capture. (d)-(f) Images from two captures and a subtraction. (g) FP reconstructions formed after capturing and processing 29 unique images (magnitude of reconstruction shown). With the optimized LED pattern in (f), areas of infection appear as bright spots that are missing from uninfected cells.  }
\label{cells}
\end{figure}


For each sample, we first capture multiple images by individually illuminating the $L=29$ uniquely positioned LEDs one at a time. We illuminate separately with the red, green and blue channel of each LED and take 3 separate images to capture color. To create our classification data set, we first identified cells with non-standard morphology and smoothness (both in bright-field and dark-field images). Then, we drew small bounding boxes around each cell and used Fourier ptychography (FP) to process each set of 29 images into an image reconstruction with higher resolution~\cite{Zheng13}. The FP reconstructions contain both the cell's absorption and phase delay at approximately 385 nm resolution (Sparrow limit), which approximately matches the quality of imaging with an oil-immersion objective lens~\cite{Ou15}. Example FP reconstructions are in Fig.~\ref{cells}(g). We then digitally refocused the FP reconstruction to the best plane of focus and visually inspected each cell to determine whether it was uninfected or contained features indicative of {\it P. falciparum} infection. After each diagnosis, we then cropped the cell with an approximate $28\times28$ pixel bounding box from each of the 29 raw images. We used cubic resizing to form $28\times28$ pixel images when necessary to make all cropped images the same size. Each $28\times28\times29$ data cube and its binary diagnosis label (infected or uninfected) form one example within our initial dataset. 

We diagnosed and extracted a total of 1258 data cubes from cells across 8 different microscope slides (563 uninfected, 695 infected). We then randomly shuffled these examples and separated them into training and testing sets. To increase algorithm robustness, we applied data augmentation (factor of 24) separately to the training and testing sets. Each final training set contained 25,000 examples and each final test set contained 5,000 examples (with 50\% uninfected and 50\% infected cells). 4 different transformations were used for data augmentation: image rotation (90 degree increments), image blurring ($3\times3$ pixel Gaussian blur kernel), elastic image deformation~\cite{Simard03}, and the addition of a small amount of Gaussian noise. These transformations align with the following experimental uncertainties that we expect in future experiments: arbitrary slide placement, slight microscope defocus, distortions caused by different smear procedures and varying amounts of sensor readout noise.

Similar to our simulation, we compare the performance of our DP-CNN's joint illumination optimization/classification process with several alternative illuminate-and-classify strategies. Average accuracies (each over 7 independent trials) and their standard deviations (STD) are presented in Table~2. Due to our light staining of each slide, direct illumination using just the center LED often produces an observable indication of cell infection, as shown in Fig.~\ref{cells}(a), top. However, accurate diagnosis at the resolution of a 20X microscope objective lens here remains challenging. As noted above, an oil-immersion objective lens is typically required for this task~\cite{WHO09}. Thus, it is not surprising that the classification accuracy here remains relatively low: approximately 90\%. Using incoherent illumination (all of the LEDs illuminate the sample simultaneously, Fig.~\ref{cells}(b)) increases the image resolution~\cite{Goodman} but significantly decreases image contrast. It appears that maintaining high contrast is extremely important for accurate classification of infection, as the classification accuracy drops to $\sim$72\% with incoherent light. Off-axis illumination (Fig.~\ref{cells}(c)) generates higher image contrast and improves accuracy to 85\%, not quite to the level of center LED illumination.

We also consider procedures that capture and subtract two images. Randomly selecting positive and negative illumination weights produces low contrast (Fig.~\ref{cells}(d)) and classification remains poor (80\%). Differential phase-contrast imaging using opposing illumination from 2 LEDs and an image subtraction does slightly better (Fig.~\ref{cells}(e), 85\%). The accuracy of our DP-CNN optimized illumination pattern, at approximately 94\%, remains significantly higher than all alternatives (Fig.~\ref{cells}(f)). In Table 2, we also present classification accuracies for each strategy based on a ensemble method over 7 independent trials (a majority vote~\cite{Lee15, Lakshminarayanan16}, 'Majority' column). Here, we define a cell as (un)infected if it is classified as (un)infected in 4 or more of the 7 independent trials. This effectively combines the 7 classifiers to increase accuracy by 2.8\% on average, and leads to approximately 98\% accuracy for our optimized LED illumination scheme. The false negative rate from the DP-CNN tests were on average approximately 16\% higher than the false positive rate (42\% of errors were false-positives), which may be partially accounted for by the unequal number of infected and uninfected cell images in our training data. Reducing this false negative rate will be a major focus of future work.

\begin{figure}
\centering
\includegraphics[width=0.8\linewidth]{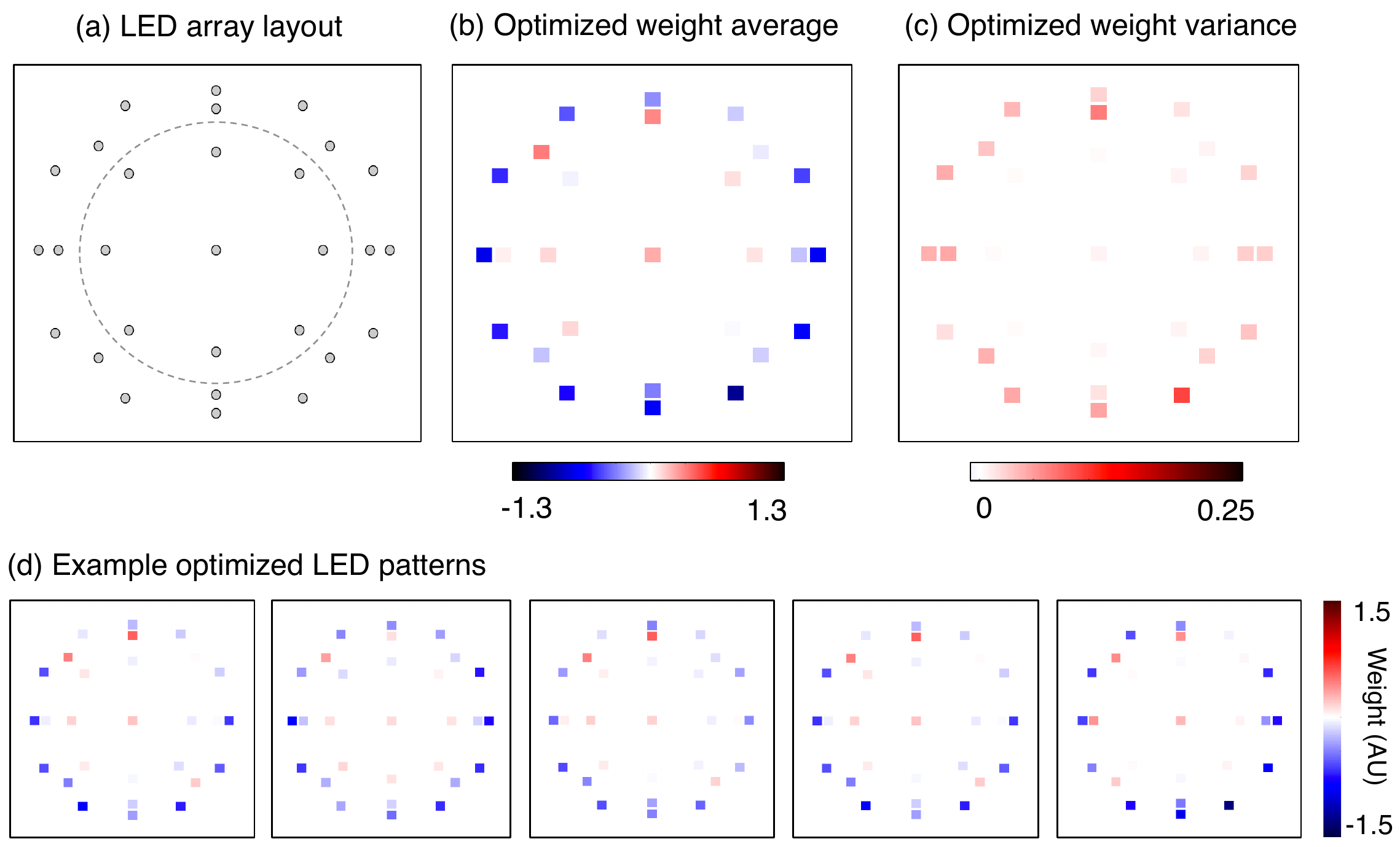}
\caption{DP-CNN optimized LED illumination patterns. (a) Location of 29 LEDs with bright/dark field boarder marked as dashed line. (b) Average LED weights and (c) variance for the illumination unit computed from 14 independent trials. (d) 5 example optimized LED patterns.}
\label{leds}
\end{figure}

Several DP-CNN optimized LED patterns are shown in Fig.~\ref{leds}. Unlike the simulation, each independent trial of DP-CNN with the experimental data converges to a similar optimal weight pattern, up to an arbitrary sign offset. That is, approximately 50\% of the optimized patterns show positive intensity weights along the outer ring and mostly negative weights within, which we can flip to match the other 50\% by multiplying all weights by -1. After accounting for this sign offset we show the average optimized LED pattern and its variance over 14 independent trials in Fig.~\ref{leds}(b)-(c). 5 example patterns from the 14 independent train/test trials are in Fig.~\ref{leds}(d). Most of the weight energy within the average pattern (75.4\%) is dedicated to the outer ring of 12 LEDs, suggesting that high spatial frequency information is important to accurately classify infection. This ring-like structure conceptually matches the condenser annulus design used in phase contrast microscopes, but here the ring is not uniformly weighted. Furthermore, 85.4\% of the average LED weight energy is negative. As less light makes it through the lens at high illumination angles, the resulting optimized images are thus noisier (see Fig.~\ref{cells}(f)). Although the classifier must compete with increased image noise, the mixture of positive LED weights from the center and negative LED weights from the outer ring visually highlight the inner-cell infections as bright spots in Fig.~\ref{cells}(f), apparently allowing the digital classifier to achieve its higher accuracy versus the alternative illumination schemes. 

\begin{table}[]
  \caption{Experimental results,  DP-CNN classification of cells with {\it P. falciparum} infection}
  \label{sample-table}
  \centering
  \begin{tabular}{lllllllll}
    \toprule
   \multicolumn{2}{c}{ }     & \multicolumn{7}{c}{Illumination Type \& Classification Score}               \\
    \cmidrule{3-9}
    Algo. (step) & Value & Center & All & Off-axis & Random & DPC & Optim. & FP   \\
    \midrule
    Adam ($10^{-4}$) & Average & 0.896 & 0.746 & 0.832 & 0.813 & 0.859 & {\bf 0.932} & 0.951   \\
     & Majority & 0.902 & 0.757 & 0.838 & 0.893 & 0.875 & {\bf 0.957} & 0.989 \\
     & STD & 0.012 & 0.012 & 0.010 & 0.037 & 0.006 & 0.019 & 0.011 \\
    \midrule
    Adam ($10^{-3}$) & Average & 0.906 & 0.689 & 0.863 & 0.806 & 0.848 & {\bf 0.949} & 0.968 \\
     & Majority & 0.920 & 0.700 & 0.886 & 0.885 & 0.881 & {\bf 0.983} & 0.987 \\
      & STD & 0.011 & 0.025 & 0.017 & 0.080 & 0.008 & 0.020 & 0.008 \\

    \bottomrule
  \end{tabular}
\end{table}

Finally, we compare the performance of DP-CNN (using 2 images) to two alternative classification methods that we expect to exhibit higher accuracy. First, we attempt to classify malaria parasite infection from the high-resolution Fourier ptychographic reconstructions of each cell. For each data cube, we use the FP algorithm~\cite{Zheng13} to process all 29 individual images to form a high-resolution complex-valued reconstruction ($56\times56$ pixels, 385 nm Sparrow limit, Fig.~\ref{cells}(g)). We apply data augmentation (same parameters) to the magnitudes of FP reconstructions to create a training/testing set with 25K/5K high-resolution labeled examples. We then process this set with the DP-CNN post-processing layers and obtain a slightly higher classification accuracy than our 2-image DP-CNN: 96\% on average and 99\% majority vote. This slight improvement is not surprising, given that FP requires a 15X increase in the amount of data that must be recorded per cell. 

Second, we obtained a baseline classification accuracy from a trained human expert who was not involved in the sample preparation or imaging experiments. After randomly selecting 50 labeled cells from our full dataset (0.5 probability uninfected, 0.5 infected), we displayed 4 images of each cell to the expert: a low-resolution image from the center LED, from the bright-field off-axis LED, from the dark-field off-axis LED, and the FP-reconstructed magnitude. Based on these 4 images, the expert made a diagnosis that matched the assigned label with an accuracy of 96\% (48 matched, 2 mismatched, both mismatches false positives). Diagnosis errors can be partially attributed to the low resolution of each raw image and artifacts present in the FP reconstruction.

\section{Discussion and future work}

In summary, we have presented a framework to jointly optimize the physical parameters of a microscope and the CNN weights used to classify the images that it generates. As a demonstration, we improved the accuracy of malaria classification by determining an optimal LED-based microscope illumination pattern. Our experiment shows that the physical parameters of a microscope are important to a particular image classification task. This should not necessarily come as a surprise, given that an optimized microscope design can capture more relevant information than a standard system (e.g., by introducing phase contrast and emphasizing important spatial frequencies).

In a number of respects, our experiments were very much a proof-of-concept and much future work is needed for a more robust malaria detection process. First, we must increase the size of our training sets with more annotated samples from a large number of patients. Second, we will involve several trained professionals to label these datasets of stained and unstained {\it P. falciparum}-infected cells at different stages of infection. To make the test as secure and reliable as possible, a major focus will be on the elimination of the false determinations, in particular the false negatives. We will also use a sorting technique (as in Ref.~\cite{Park16}) to first separate infected from uninfected cells before imaging to obtain a better ground-truth classification. Finally, we need to compare the performance of our CNN pipeline with prior methods for malaria classification that use standard microscope images~\cite{Das13,Quinn16}. A combination of our physical optimization process with these alternative learning pipelines may yield a higher overall classification accuracy. In general, we anticipate that this technology may be directly extended to diagnosis {\it P. falciparum} infection at a per patient level (as opposed to a per-cell level). Given the larger number of infected cells per smear, we expect much higher patient-specific classification accuracies as opposed to the per-cell classification accuracies reported here.

Outside of the particular goal of classifying malaria infection, we see a number of potential extensions of our new framework. First, our optimized LED patterns require two images (assuming positive and negative weights). Following recent work with spectral classification~\cite{Dunlop16}, a future strategy may use an adaptive approach to classify with more than two uniquely lit images (e.g., the LED pattern for each subsequent image can depend on the classification scores for previous images). Second, there are a large number of alternative physical aspects of a microscope (or an imaging system in general) to potentially optimize via a CNN. An immediate extension could jointly optimize both the illumination and a phase plate (aperture mask) to place in the microscope back focal plane, which effectively customizes the condenser annulus and phase ring in phase contrast setups for a particular CNN-based goal. Other microscope design options include (but are not limited to) optimizing spectral and polarimetric degrees of freedom, improving axial resolution and depth detection, addressing various sensor characteristics, and enhancing abilities to see through scattering material like tissue. Many of these aspects may find their way into the physical layers of our P-CNN model. In general, we are hopeful that the initial foundation presented here will help others explore more connections between artificial neural networks and the physical devices used to acquire their data.

\subsubsection*{Acknowledgments}
We thank Xiaoze Ou, Jaebum Chung and Prof.\ Changhuei Yang for capturing and sharing the raw images used in the experiment and thank the lab of Prof.\ Ana Rodriguez for providing the malaria-infected cells. R.H. acknowledges funding from the Einstein Foundation Berlin.  

\section*{Appendix A: Algorithm details}
All results were obtained using Tensorflow. For both the simulations and experiments, we used the cross-entropy error metric and also applied dropout before the final readout layer. For our simulation results, we trained with 50,000 iterations and a batch size of 50 and used the Adam method for stochastic optimization~\cite{Kingma14} with a step size of $10^{-4}$. For the experimental results, we trained with 25,000 iterations using a batch size of 50 and report results using Adam Optimization with a step size of $10^{-3}$ and $10^{-4}$. For the Adam optimizer we set $\beta_1=0.9$, $\beta_2=0.999$ and $\epsilon=$1e-8. Here is a brief review of the experimental DP-CNN pipeline in Fig.~\ref{cnnmodel}(c): layer 1 (physical layer) is a tensor product between a $28\times28\times29$ input and a $1\times29$ weight vector $w$, layer 2 uses a $5\times5$ convolution kernel, 32 features, ReLU activation and $2\times2$ max pooling, layer 3 uses a $5\times5$ convolution kernel, 64 features, ReLU activation and $2\times2$ max pooling, layer 4 is a densely connected layer with 1024 neurons and ReLU activation and layer 5 is densely connected readout layer. We apply dropout with a probability of 0.5 between layers 4 and 5.

\end{document}